\documentclass[runningheads]{llncs}
\pdfoutput=1
\usepackage[T1]{fontenc}
\usepackage{graphicx}

\usepackage{amsmath}
\usepackage{amssymb}
\usepackage{mathrsfs}

\usepackage{algorithm}
\usepackage{algorithmicx}
\usepackage{algpseudocode}
\usepackage{booktabs}
\usepackage{bbding}
\usepackage{cite}

\begin{document}
\title{Locality Constrained Analysis Dictionary Learning via K-SVD Algorithm}
\titlerunning{Locality Constrained Analysis Dictionary Learning}
%
\author{Kun Jiang\inst{(}\Envelope\inst{)} \and
Zhaoli Liu\inst{}\and
Zheng Liu\inst{}\and
Qindong Sun\inst{}}
\authorrunning{K. Jiang et al.}
%
\institute{School of Computer Science and Engineering\\Xi'an University of Technology, Xi'an, China\\
\email{ jk\_365@126.com}\\
\email{\{zhaoliliu,zhengliu,sqd\}@xaut.edu.cn}}
\maketitle              
\begin{abstract}
Recent years, analysis dictionary learning (ADL) and its applications for classification have been well developed, due to its flexible projective ability and low classification complexity. With the learned analysis dictionary, test samples can be transformed into a sparse subspace for classification efficiently. However, the underling locality of sample data has rarely been explored in analysis dictionary to enhance the discriminative capability of the classifier. In this paper, we propose a novel locality constrained analysis dictionary learning model with a synthesis K-SVD algorithm (SK-LADL). It considers the intrinsic geometric properties by imposing graph regularization to uncover the geometric structure for the image data. Through the learned analysis dictionary, we transform the image to a new and compact space where the manifold assumption can be further guaranteed. thus, the local geometrical structure of images can be preserved in sparse representation coefficients. Moreover, the SK-LADL model is iteratively solved by the synthesis K-SVD and gradient technique. Experimental results on image classification validate the performance superiority of our SK-LADL model.

\end{abstract}
\section{Introduction}
Sparse representation which encodes the signals or images using only a few active coefficients, has been successfully applied to many areas across signal processing and pattern recognition. Signals can be represented as a linear combination of a relatively small number of atoms in an over-complete dictionary. Sparse representation models can be classified into two categories by the learning method, synthesis dictionary learning (SDL) and analysis dictionary learning (ADL) \cite{1710377, 6339105}. A typical SDL model expects to learn the over-completed dictionary by minimizing the reconstruction errors such that it can linearly represent the original signals. Instead of learning an over-complete representation dictionary in SDL, the ADL model mainly focuses on learning a transformation matrix, and constructing sparse analyzed vectors. The ADL model has aroused much attention as it has a more intuitive illustration for the role of analysis atoms and high efficiency.

The traditional SDL model views the signal reconstruction problem as a pure approximation task, which may overlook signal intrinsic attributes, such as signal structure and density. To overcome this issue, several sparse coding models incorporating the geometrical structures of the image space have been proposed. They are based on the locally invariant idea, which assumes that two close points in the original space are likely to have similar encodings. Thus, additional regularizers are incorporated in order to satisfy specific requirements in various application scenarios. For example, graph regularized sparse coding based on nonlinear manifold learning is proposed for image classing and clustering applications \cite{8467342,6572465,6459595,8920031}. Motivated by recent progress in sparse representation, we combine the ADL model with manifold learning, called locality constrained ADL with a synthesis K-SVD solver (SK-LADL). The flowchart of our proposed method for classification is illustrated in Fig. \ref{1}. The novel strategy incorporates the graph regularizer to emphasize the correlations between neighbors of training data, which could enhance the discriminative ability of the analysis dictionary.

\begin{figure}
\centering
\includegraphics[width=0.5\textwidth]{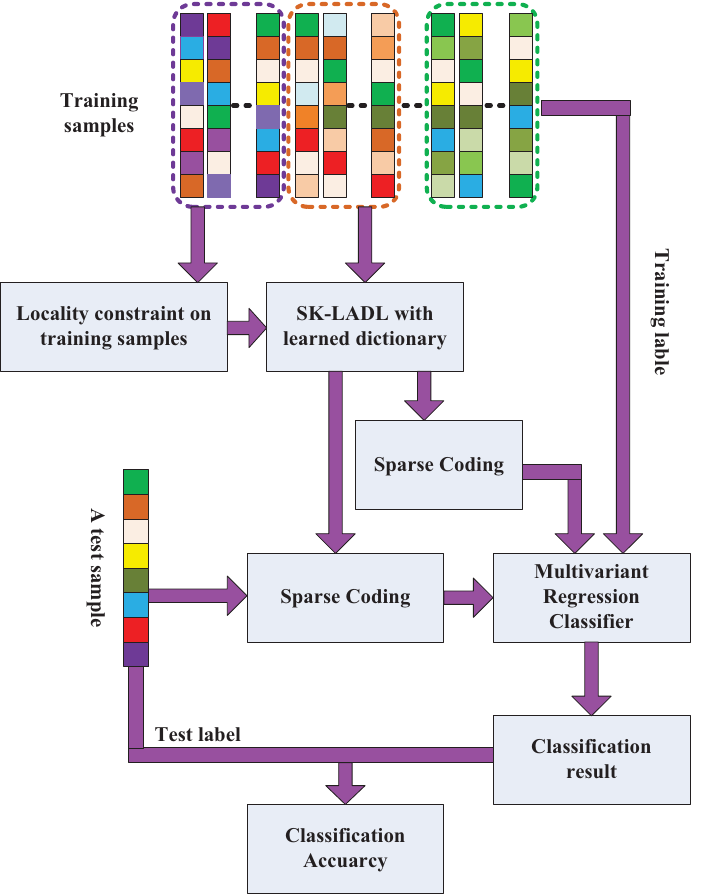}
\caption{Flowchart of our proposed SK-LADL model for classification.} \label{fig1}
\end{figure}

The rest of this paper is organized as follows. The basic dictionary learning framework is presented in Section 2. The locality constrained analysis dictionary learning model is presented in Section 3. The K-SVD algorithm is introduced in Section 4. The classification methodology adopted in our model is reviewed in Section 5. Experimental results on image classification are presented in Section 6. Relevant conclusions are finally given in Section 7.

\section{Background}

\subsection{Synthesis Dictionary Learning}
Given signals $Y=[y_1,y_2,\cdots,y_n]\in R^{m\times n}$, let $D=[d_1,d_2,\cdots,d_k]\in R^{m\times k}$ be a synthesis dictionary with a serials of atom $d_i$, and $X=[x_1,x_2,\cdots,x_n]\in R^{k\times n}$ be the sparse coefficient matrix. The main idea of synthesis dictionary learning (SDL) is to approximately reconstruct the original signals $Y$ by the combination of dictionary atoms $d_i(1\leq i \leq k)$ with respective weight factors or coefficients $X$. The sparse optimization problem of SDL can be formulated as follows.
\begin{equation}\label{1}
  \min_{D,X} (\|Y-DX\|_F^2+\lambda\sum^M_{i=1}\|x_i\|_0)
\end{equation}
where $\|Y-DX\|_F^2$ is the reconstruction error term and $\lambda$ is a positive weight coefficient \cite{1710377}.

Sparse representation of the above SDL model can be classified in two basic tasks, i.e., sparse approximation and dictionary learning. The task of sparse approximation is to find a sparse solution to an underdetermined linear system. Some algorithms, such as matching pursuit (MP), basis pursuit (BP) and shrinkage method, have been well developed. On the other hand, dictionary learning is dedicated to search an optimal signal space to support the attribution of sparse vector under a certain measure. There exist a variety of numerical algorithms presented to achieve this objective, e.g., the method of optimal directions (MOD), K-means singular value decomposition (K-SVD) and recursive least squares (RLS).

The sparse analysis dictionary learning (ADL) model aims to learn a projective matrix (i.e., analysis dictionary) $\Omega \in R^{k\times m}$ with $k>m$ to implement the approximately sparse representation of the signal $y\in R^m$ in transformed domain\cite{6339105,6339108}. Specifically, it assumes that the product of $\Omega$ and $y$ is sparse, i.e., $x=\Omega y$ with $\|x_i\|_0=k-l$, where $0\leq l\leq k$ is the number of zeros in $x \in R^k$. The sparse optimization problem can be formulated as follows.
\begin{equation}\label{2}
  \min_{\Omega,X} (\|X-\Omega Y\|_F^2+\lambda\sum^M_{i=1}\|x_i\|_0)
\end{equation}
where the representation error term $\|X-\Omega Y\|_F^2$ shows the disparity between image representations in the transformed space and the coefficients with target sparsity level.
The optimization problem in equation (2) above can be rewritten in the vector form as
\begin{equation}\label{3}
 \min_{\Omega,\{x_i\}} (\|\Omega y_i-x_i\|_2^2+\lambda\|x_i\|_0)
\end{equation}
If the signal $y_i$ is known, its analysis representation with respect to a given dictionary can be obtained via multiplying by $\Omega$. However, when the observed signal is contaminated by noise, the clean signal has to be estimated first in order to get its analysis representation, which leads to the analysis pursuit problem. Some algorithms like backward-greedy (BG), optimized-backward-greedy (OBG), and greedy analysis pursuit (GAP) have been proposed to address this problem \cite{6339105}. The computational complexity of the basic ADL model is very high, and some recent work has relax the  $l_0$-norm in equation (3) to  $l_1$-norm or thresholding function\cite{7026054}.

\section{Locality Constrained Analysis Model}
The local geometric structure such as texture and contour information can be added to the sparse representation problem to upgrade the accuracy of signals approximation. Locality is more essential than sparsity, since locality leads to sparsity but not necessary vice versa \cite{NIPS2009_2afe4567}. The manifold learning theory can be used to quantify intrinsic geometric locality structure information. Many previous works have demonstrated that local structure embedded in high-dimensional vector space is very important to characterize both the intrinsic structure and discriminative structure of images \cite{GAO20152543}. Moreover, since the data is more likely to reside on a low-dimensional sub-manifold embedded in the high-dimensional ambient space, the geometrical information of the data is important for discrimination \cite{8467342,7365468,8920031}. Therefore, by preserving the locality characteristics of the training samples, the discriminative ability of the learned dictionary can be improved.

Motivated by previous works, a natural assumption here is that if two samples $y_i$ and $y_j$ are close to each other in the intrinsic geometric structure of the data distribution, the sparse representation vectors $x_i$ and $x_j$ over dictionary $\Omega$ , should be also close to each other. By use of techniques from Laplacian eigenmap method, we employ the graph Laplacian to build the local manifold structure. Specifically, we construct a supervised nearest neighbor graph $\mathbb{G}=\{Y, W\}$ with $N$ vertices, in which each vertex denotes a data point, and the similarity matrix $W$ over training samples is defined as
\begin{equation}\label{4}
  {{W}_{ij}}=\left\{ \begin{matrix}
   \exp \left( -{{\left\| {{y}_{i}}-{{y}_{j}} \right\|}_{2}}/\delta  \right)  \\
   0  \\
\end{matrix} \right.\begin{matrix}
   {} & \begin{matrix}
   if\ y_j \in \mathbb{N}(y_i)\\
   otherwise  \\
\end{matrix}  \\
\end{matrix}
\end{equation}
where $\delta$ is kernel width and $\mathbb{N}(y_i)$ is the $k$-nearest neighbors of sample $y_i$.
To map the supervised nearest neighbor graph $\mathbb{G}$ to the sparse representation matrix $X$ we minimize the following graph Laplacian criterion
\begin{equation}\label{5}
\begin{aligned}
  \frac{1}{2}\sum\limits_{i=1}^{N}{\left\| {{x}_{i}}-{{x}_{j}} \right\|}_{2}^{2}{{W}_{i,j}}& =\sum\limits_{i=1}^{N}{{{x}_{i}}^{T}{{x}_{i}}{{B}_{ii}}}-\sum\limits_{i=1}^{N}{{{x}_{i}}^{T}{{x}_{i}}{{W}_{ij}}} \\
 & =Tr\left( XB{{X}^{T}} \right)-Tr\left( XW{{X}^{T}} \right) \\
 & =Tr\left( XL{{X}^{T}} \right) \\
\end{aligned}
\end{equation}
where $L=B-W$ is called as graph Laplacian matrix, $B$ is a diagonal weight matrix with $i$th diagonal entry is ${{B}_{ii}}=\sum{{{W}_{ji}}}$.

Thus, with the regularizer $Tr(XLX^T)$, the corresponding representations of any two points $y_i$ and $y_j$ are expected to maintain the same local structure of them. Then the sparse representation problem can be formulated as
\begin{equation}\label{6}
  \underset{\Omega ,X}{\mathop{\min }}\,\left( \left\| X-\Omega Y \right\|_{F}^{2}+\lambda \sum\limits_{i=1}^{M}{{{\left\| {{x}_{i}} \right\|}_{0}}}+\alpha Tr\left( XL{{X}^{T}} \right) \right)
\end{equation}
where $\alpha$ is weight coefficient, $L$ is a weight matrix depended on the chosen manifold. As the Laplacian is symmetric and positive semidefinite, the objective (6) is a convex optimization problem. We rewrite the Laplacian regularizer $Tr( XL{{X}^{T}})$ in equation (6) to the vector form as
\begin{equation}\label{7}
 Tr\left( XL{{X}^{T}} \right)=Tr\left( \sum\limits_{i,j=1}^{M}{{{L}_{i,j}}{{x}_{i}}x_{j}^{T}} \right)=\sum\limits_{i,j=1}^{M}{{{L}_{i,j}}x_{i}^{T}{{x}_{j}}}
\end{equation}

The optimization problem in equation (6) above can be rewritten in the vector form as
\begin{equation}\label{8}
  \underset{\Omega ,\left\{ {{x}_{i}} \right\}}{\mathop{\min }}\,\sum\limits_{i=1}^{M}{\left( \left\| \Omega {{y}_{i}}-{{x}_{i}} \right\|_{2}^{2}+\lambda {{\left\| {{x}_{i}} \right\|}_{0}}+\alpha \sum\limits_{j=1}^{M}{{{L}_{i,j}}x_{i}^{T}{{x}_{j}}} \right)}
\end{equation}

When the dictionary $\Omega$ and all the other vectors ${{\left\{ {{x}_{k}} \right\}}_{k\ne i}}$ are fixed, we obtain the following optimization problem
\begin{equation}\label{9}
  \underset{{{x}_{i}}}{\mathop{\min }}\,\left\| \Omega {{y}_{i}}-{{x}_{i}} \right\|_{2}^{2}+\lambda {{\left\| {{x}_{i}} \right\|}_{0}}+\alpha \sum\limits_{j=1}^{M}{{{L}_{i,j}}x_{i}^{T}{{x}_{j}}}
\end{equation}

After some manipulations, problem (9) can be further cast as
\begin{equation}\label{10}
  \underset{{{x}_{i}}}{\mathop{\min }}\,\left\| \Omega {{y}_{i}}-{{x}_{i}} \right\|_{2}^{2}+\lambda {{\left\| {{x}_{i}} \right\|}_{0}}+\alpha {{L}_{ii}}\left\| {{x}_{i}}+\left( 1/\left( 2{{L}_{ii}} \right)\sum\limits_{k\ne i}{{{L}_{ki}}{{x}_{k}}} \right) \right\|_{2}^{2}
\end{equation}

Then, merging two  $l_2$-norm terms in (10) yields the final SK-LADL model as
\begin{equation}\label{11}
  \underset{{{x}_{i}}}{\mathop{\min }}\,\left\| \left( \begin{matrix}
   \Omega {{y}_{i}}  \\
   -\sqrt{\alpha {{L}_{ii}}}X{{p}_{i}}  \\
\end{matrix} \right)-\left( \begin{matrix}
   R  \\
   \sqrt{\alpha {{L}_{ii}}}I  \\
\end{matrix} \right){{x}_{i}} \right\|_{2}^{2}+\lambda {{\left\| {{x}_{i}} \right\|}_{0}}
\end{equation}
where $I$ is an identity matrix and $R$ is initialized as an identity matrix, each element of $p_i$ is defined by
\begin{equation}\label{12}
  {{p}_{ik}}=\left\{ \begin{matrix}
   {{L}_{ik}}/\left( 2{{L}_{ii}} \right)  \\
   0  \\
\end{matrix} \right.\begin{matrix}
   {} & \begin{matrix}
   i\ne k  \\
   i=k  \\
\end{matrix}  \\
\end{matrix}
\end{equation}
\section{Solving the Problem}
In this section, we adopt an alternative strategy to solve the SK-LADL model. The synthesis K-SVD and gradient technique are incorporated into the optimization procedure. The iterative optimization algorithm contains the following two steps:

\textbf{Update} $\{X\}$ Fixing the dictionary $\Omega$, the solutions for $X$ in the present iteration can be obtained by utilizing the synthesis K-SVD. Denote the two combined matrices in the parentheses of problem (11) as
\begin{equation}\label{13}
  {{y}_{ne{{w}_{i}}}}={{\left( {{y}_{i}}^{T}{{\Omega }^{T}},-\sqrt{\alpha {{L}_{ii}}}p_{i}^{T}{{X}^{T}} \right)}^{T}}
\end{equation}

\begin{equation}\label{14}
  {{Q}_{new}}={{\left( {{R}^{T}},\sqrt{\alpha {{L}_{ii}}}I \right)}^{T}}
\end{equation}

In the original K-SVD algorithm, the matrix $Q_{new}$ is column-wise $l_2$ normalized. The optimization problem (11) is equivalent to the following problem:
\begin{equation}\label{15}
  \begin{matrix}
   \left\langle {{Q}_{new}},X \right\rangle =\underset{{{Q}_{new}},X}{\mathop{\arg \min \left\| {{y}_{new}}-{{Q}_{new}}{{x}_{i}} \right\|}}\,_{2}^{2}  \\
   \begin{matrix}
   s.t. & {{\left\| {{x}_{i}} \right\|}_{0}}\le {{T}_{0}},\forall i=1,\cdots ,n.  \\
\end{matrix}  \\
\end{matrix}
\end{equation}

The optimization problem in (15) can be solved by the K-SVD method, with entrance parameters being $y_{new}$, $Q_{new}$ and $T_0$. Let $x_R^k$ be the corresponding coefficients of the $k$th column of $Q_{new}$, and denoted as $q_k$. Let ${{E}_{k}}=\left( {{Y}_{new}}-\sum\nolimits_{j\ne k}{{{q}_{j}}x_{R}^{j}} \right)$, then discard the zero entries in $x_R^k$ and $E_k$, with corresponding results marked as $\tilde{x}_R^k$ and $\tilde{E}_k$ respectively. We optimize the following problem to obtain $q_k$ and $\tilde{x}_R^k$.
\begin{equation}\label{16}
  \left\langle {{q}_{k}},\tilde{x}_{R}^{k} \right\rangle =\underset{{{q}_{k}},\tilde{x}_{R}^{k}}{\mathop{\arg \min }}\,\left\| {{{\tilde{E}}}_{k}}-{{q}_{k}}\tilde{x}_{R}^{k} \right\|
\end{equation}

Decomposing $\tilde{E}_k$ by K-SVD method, we have ${{\tilde{E}}_{k}}=U\Sigma {{V}^{T}}$. Let $\tilde{x}_{R}^{k}=\Sigma \left( 1,1 \right)V\left( :,1 \right)$. After that, the nonzero values of $x_R^k$ are replaced by $\tilde{x}_R^k$. Then the dictionary $R^{t}$ in $Q_{new}^{t}$ can be used for the next iterator, i.e., $R^{t+1}=R^{t}$. And the $Q_{new}^{t+1}$ is updated accordingly with $R^{t+1}$ and the Laplacian matrix $L$.

\textbf{Update} $\{\Omega\}$ Fixing $X$, the solution for $\Omega$ is computed through matrix derivation, followed by normalization of the rows of $\Omega$. We add the following regularization term into the object function. The formulation in this situation is given as
\begin{equation}\label{17}
  \underset{{{x}_{i}}}{\mathop{\min }}\,\left\| \Omega {{y}_{i}}-{{x}_{i}} \right\|_{2}^{2}+\alpha {{L}_{ii}}\left\| {{x}_{i}}+\left( 1/\left( 2{{L}_{ii}} \right)\sum\limits_{k\ne i}{{{L}_{ki}}{{x}_{k}}} \right) \right\|_{2}^{2}+\beta \left\| \Omega  \right\|_{2}^{2}
\end{equation}
After omitting the independent terms, an equivalent problem is obtained as follows.
\begin{equation}\label{18}
  \min \left( y_{i}^{T}{{\Omega }^{T}}\Omega {{y}_{i}}-2{{x}_{i}}^{T}\Omega {{y}_{i}}+\beta {{\Omega }^{T}}\Omega  \right)
\end{equation}
where $\beta$ is a parameter which indicates the weight of the penalty term. The role of this term lies in avoiding singularity and over-fitting issues as well as ensuring the stable solution of dictionary.

Let the first derivative w.r.t. $\Omega$ be zero and we can obtain the closed-form analytical solution for the dictionary as
\begin{equation}\label{19}
  \Omega ={{x}_{i}}{{y}_{i}}^{T}{{\left( {{y}_{i}}{{y}_{i}}^{T}+\beta I \right)}^{-1}}
\end{equation}

Then we normalize rows of the dictionary, which results in better performance empirically. The above training procedure is outlined in Algorithm 1.


\floatname{algorithm}{Algorithm}
\renewcommand{\algorithmicrequire}{\textbf{Input:}}
\renewcommand{\algorithmicensure}{\textbf{Output:}}
\begin{algorithm}
\caption{Algorithm for solving our proposed model}
\begin{algorithmic}[1]
\Require Training data $Y$, model parameter $\alpha$,$\beta$.
\Ensure The analysis dictionary $\Omega$.
\State Initialize $\Omega$ as random matrix, calculate the Laplacian matrix $L$;
\While{not converge}
\State fixing $\Omega$, update $X$ by solving model (15) using synthesis K-SVD;
\State fixing $X$, update $\Omega$ by Eq. (19);
\State check the convergence condition:
\State ${{\left\| {{X}^{t}}-{{\Omega }^{t}}Y \right\|}_{\infty }}<\varepsilon $;
\State $t=t+1$;
\EndWhile
\end{algorithmic}
\end{algorithm}
\section{Classification Methodology}
For classification of test samples on SK-LADL model, preliminary representation coefficient $x_i$ are obtained via the operation of multiplying $\Omega$ by the testing sample $y_i$. Keeping in view that the proposed SK-LADL method imposes sparsity on coefficients, a hard thresholding operator is used to maintain the sparse characteristic of $x_i$. The operator reserves elements with $T_0$ biggest absolute values and sets the others to be zero. Then, the multivariant ridge regression model is used to obtain a linear classifier from the sparse representations $X$ for training samples $Y$:
\begin{equation}\label{20}
  \underset{W}{\mathop{\min }}\,\left\| H-WX \right\|_{F}^{2}+\delta \left\| W \right\|_{F}^{2}
\end{equation}
where $\delta$ is a parameter for regularizing the solution, $H=\left[ {{h}_{1}},{{h}_{2}},\cdots ,{{h}_{n}} \right]$ is the label matrix of sparse analysis representation $X$, and each column ${{h}_{i}}={{\left[ 0,\cdots ,0,1,0,\cdots ,0 \right]}^{T}}$ describes the label vector of the $i$th sample. The optimal solution of above problem is
\begin{equation}\label{21}
  {{W}^{*}}=H{{X}^{T}}{{\left( H{{X}^{T}}+\delta I \right)}^{-1}}
\end{equation}

With the optimized $W^*$, a testing sample $y_i$ can be predicted by picking the index of the maximum element of $W^*x_i$. Then let the predictive vector ${{h}_{{{i}_{pre}}}}=W{{x}_{i}}$. The predictive vector ${{h}_{{{i}_{pre}}}}$ has a very approximate shape to the corresponding real label vector $h_i$, with only one element obviously bigger than others. Therefore, the location of the largest element in ${{h}_{{{i}_{pre}}}}$ is utilized to determine the category.
\section{Experiments}
\subsection{Experiment Setups}
In this section, we evaluate our SK-LADL model on four public image datasets, Extended YaleB (EYaleB), AR, Scene15 and UCF50. The above databases are widely used in evaluating the performance of sparse representation-based classification methods. The features are provided by \cite{6516503} and \cite{6247806}. On EYaleB and AR datasets, random features are generated by the projection with a randomly generated matrix. On Scene15 datasets, features are achieved by extracting SIFT descriptors, max pooling in spatial pyramid and reducing dimensions by PCA. UCF50 is a large-scale and challenging action recognition database. It has 50 action categories and 6680 realistic human action videos collected from YouTube. AR face database contains illumination, expression, and occlusions variations. We choose a subset consisting of 2600 face images from 50 males and 50 females.

We compare our model with some state-of-the-art approaches: SRC\cite{4483511}, K-SVD\cite{1710377}, D-KSVD\cite{5539989}, LC-KSVD\cite{6516503}, ADL-SVM\cite{7026054} and SK-DADL\cite{10.1007/s11042-017-5269-6} with the above benckmark image datasets features. For fair comparison, the experiment settings we follow are in accordance with \cite{6516503} and \cite{10.1007/s11042-017-5269-6}. The sparsity is set as 45 in all the methods, and the dictionary atom is set between 500 and 600 which is the integral multiple of the number of classes in different datasets. There are four parameters in SK-LADL model, i.e., $\alpha$, $\beta$, $\delta$ and $k$, where $\delta=0.01$ and $k=3$ are preset and the parameter $\alpha$ and $\beta$ are tuned by 5-fold cross validation and optimized by using grid search strategy. We firstly search in the larger range of $[10^{-3}, 10^{-2}, \cdots, 10^2, 10^3]$ for each parameter and then search a smaller grid with proper interval size determined by preliminary classification results.The best parameters we set in each database are listed in Table 1.

\begin{table}
\centering
\caption{Parameter selection in the best performance for parameter $\alpha$ and $\beta$.}\label{tab1}
\setlength{\tabcolsep}{4mm}
\begin{tabular}{ccccc}
\toprule
\quad\quad&EYaleB&	\quad AR\quad&	Scene15&	UCF50\\
\midrule
$\alpha$ &10&	\quad 10\quad&	20&	10\\
$\beta$ &0.03&	\quad 0.05\quad&	0.11&	0.01\\
\bottomrule
\end{tabular}
\end{table}

\subsection{Results and Analyses}
We repeat the experiments 5 times on different selected training and testing image features, and the mean accuracies are reported. Table 2 shows the mean classification accuracy results on different datasets. As can be seen, our method achieves notably higher accuracy than SRC, KSVD and LC-KSVD on all four databases. This is mainly due to the locality preservation achieved in our method for ADL model, which ensures that similar training samples tend to have similar coding coefficients. The SRC model that directly uses all training samples as the dictionary will introduce noise for the sparse representation. The locality constraint on representation coefficients can narrow the selection of representative analysis atoms on geometric manifold, enhance the representational ability of homogeneous samples to some extent, and this may help to overcome the above noise disadvantage.

Our method also achieves favorable results compared with the two ADL-based methods. This is mainly because the ADL-SVM and SK-DADL model only utilize joint- or post-learned classifiers with analysis representation without inherit the underlining structure of training samples. Therefore, with only a simple post-learned classifier, the integration of the locality information yields attractive discrimination for SK-LADL model compared to other conventional ADL models.

\begin{table}
\centering
\caption{Classification accuracy (\%) comparison on different datasets. }\label{tab2}
\setlength{\tabcolsep}{2.4mm}
\begin{tabular}{ccccccc}
\toprule
&SRC&	K-SVD&	LC-KSVD&	ADL-SVM&	SK-DADL&	SK-LADL\\
\midrule
EYaleB&	96.5&	93.1&	96.7&	95.4&	96.7&	96.4\\
AR&	97.5&	86.5&	97.8&	96.1&	97.7&	97.2\\
Scene15&	91.8&	86.7&	92.9&	91.8&	97.4&	98.2\\
UCF50&	68.4&	51.5&	70.1&	72.3&	74.6&	75.2\\
\bottomrule
\end{tabular}
\end{table}

As for the testing efficiency, Table 3 shows the time for classifying one testing image on databases EYaleB (dictionary size = 570) and AR (dictionary size = 600). As can be seen, our method performs better than SK-DADL, due to the locality constraint on sparse coding, which could generate similar analysis representation of heterogeneous samples. The accuracies and time costs in tables can demonstrate that our SK-LADL model has huge potential in pattern classification tasks.

\begin{table}
\centering
\caption{The time (ms) for classifying one testing image.}\label{tab3}
\setlength{\tabcolsep}{5mm}
\begin{tabular}{ccc}
\toprule
&EYaleB&	AR\\
\midrule
SK-DADL&	0.029&	0.078\\
SK-LADL&	0.028&	0.073\\
\bottomrule
\end{tabular}
\end{table}

\begin{figure}
\includegraphics[width=1.0\textwidth]{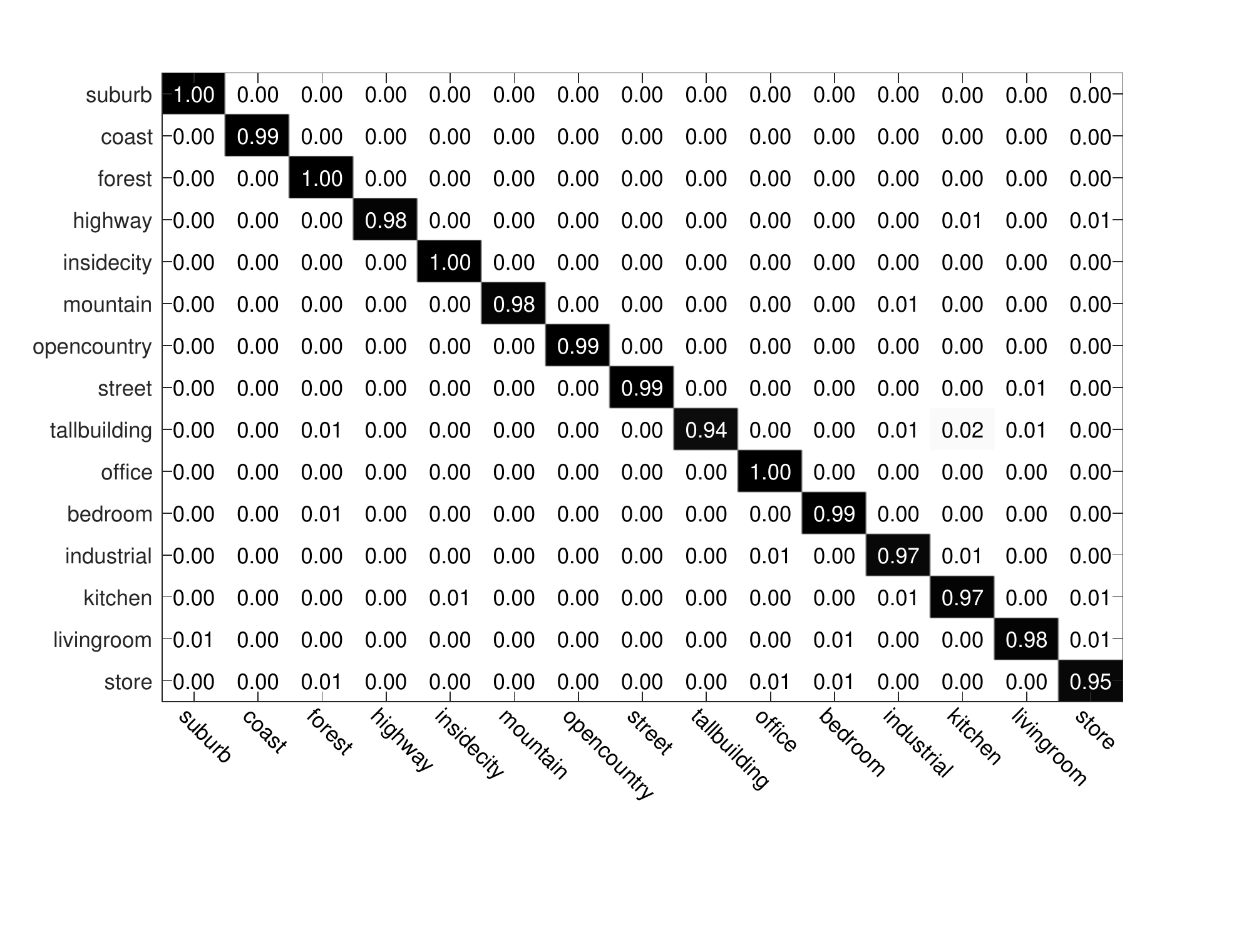}
\caption{ Confusion matrix of the ground truth on Scene 15 dataset.} \label{fig3}
\end{figure}

Fig. \ref{fig3} shows the confusion matrix for our proposed SK-LADL method on Scene15 dataset. It presents proportion of images in each category classified to all categories. We can observe that most images can be classified into the right category, with some class even getting all right classification. From the figures, we can conclude that the desired effect of our proposed SK-LADL method is reached.

\section{Conclusion}
In this paper, we proposed a novel discriminative analysis dictionary learning model with locality constrained regularizer(SK-LADL), which takes into account the intrinsic geometric structure of the training samples by introducing the locality constraint term into the framework. To effectively solve the proposed objective function, an iterative algorithm based on the conventional synthesis K-SVD and gradient technique are applied to our SK-LADL model. Experimental results on four benchmark image datasets classification demonstrate the superiority of the SK-LADL method over other state-of-the-art ADL approaches. In the future, we will incorporate locality constraint on dictionary atom and coefficients, addressing the problem when both training and test images are corrupted.
\section*{Acknowledgement}
This work is supported by the Natural Science Basic Research Plan in Shaanxi Province of China (Grant No. 2021JM-339, 2020JQ-647).
%
%
%
\bibliographystyle{splncs04}
\bibliography{mybibliography}

\begin{thebibliography}{10}
\providecommand{\url}[1]{\texttt{#1}}
\providecommand{\urlprefix}{URL }
\providecommand{\doi}[1]{https://doi.org/#1}

\bibitem{1710377}
{Aharon}, M., {Elad}, M., {Bruckstein}, A.: K-svd: An algorithm for designing
  overcomplete dictionaries for sparse representation. IEEE Transactions on
  Signal Processing  \textbf{54}(11),  4311--4322 (2006).
  \doi{10.1109/TSP.2006.881199}

\bibitem{GAO20152543}
Gao, Q., Huang, Y., Zhang, H., Hong, X., Li, K., Wang, Y.: Discriminative
  sparsity preserving projections for image recognition. Pattern Recognition
  \textbf{48}(8),  2543--2553 (2015).
  \doi{https://doi.org/10.1016/j.patcog.2015.02.015},
  \url{https://www.sciencedirect.com/science/article/pii/S0031320315000758}

\bibitem{6459595}
{Hawe}, S., {Kleinsteuber}, M., {Diepold}, K.: Analysis operator learning and
  its application to image reconstruction. IEEE Transactions on Image
  Processing  \textbf{22}(6),  2138--2150 (2013).
  \doi{10.1109/TIP.2013.2246175}

\bibitem{8467342}
{Jiang}, J., {Yu}, Y., {Wang}, Z., {Liu}, X., {Ma}, J.: Graph-regularized
  locality-constrained joint dictionary and residual learning for face sketch
  synthesis. IEEE Transactions on Image Processing  \textbf{28}(2),  628--641
  (2019). \doi{10.1109/TIP.2018.2870936}

\bibitem{6516503}
{Jiang}, Z., {Lin}, Z., {Davis}, L.S.: Label consistent k-svd: Learning a
  discriminative dictionary for recognition. IEEE Transactions on Pattern
  Analysis and Machine Intelligence  \textbf{35}(11),  2651--2664 (2013).
  \doi{10.1109/TPAMI.2013.88}

\bibitem{7365468}
{Li}, Z., {Lai}, Z., {Xu}, Y., {Yang}, J., {Zhang}, D.: A locality-constrained
  and label embedding dictionary learning algorithm for image classification.
  IEEE Transactions on Neural Networks and Learning Systems  \textbf{28}(2),
  278--293 (2017). \doi{10.1109/TNNLS.2015.2508025}

\bibitem{6339108}
{Ravishankar}, S., {Bresler}, Y.: Learning sparsifying transforms. IEEE
  Transactions on Signal Processing  \textbf{61}(5),  1072--1086 (2013).
  \doi{10.1109/TSP.2012.2226449}

\bibitem{6339105}
{Rubinstein}, R., {Peleg}, T., {Elad}, M.: Analysis k-svd: A
  dictionary-learning algorithm for the analysis sparse model. IEEE
  Transactions on Signal Processing  \textbf{61}(3),  661--677 (2013).
  \doi{10.1109/TSP.2012.2226445}

\bibitem{6247806}
{Sadanand}, S., {Corso}, J.J.: Action bank: A high-level representation of
  activity in video. In: 2012 IEEE Conference on Computer Vision and Pattern
  Recognition. pp. 1234--1241 (2012). \doi{10.1109/CVPR.2012.6247806}

\bibitem{7026054}
{Shekhar}, S., {Patel}, V.M., {Chellappa}, R.: Analysis sparse coding models
  for image-based classification. In: 2014 IEEE International Conference on
  Image Processing (ICIP). pp. 5207--5211 (2014).
  \doi{10.1109/ICIP.2014.7026054}

\bibitem{6572465}
{Tang}, Y., {Shen}, Y., {Jiang}, A., {Xu}, N., {Zhu}, C.: Image denoising via
  graph regularized k-svd. In: 2013 IEEE International Symposium on Circuits
  and Systems (ISCAS). pp. 2820--2823 (2013). \doi{10.1109/ISCAS.2013.6572465}

\bibitem{10.1007/s11042-017-5269-6}
Wang, Q., Guo, Y., Guo, J., Kong, X.: Synthesis k-svd based analysis dictionary
  learning for pattern classification. Multimedia Tools Appl.  \textbf{77}(13),
   17023--17041 (Jul 2018). \doi{10.1007/s11042-017-5269-6},
  \url{https://doi.org/10.1007/s11042-017-5269-6}

\bibitem{4483511}
{Wright}, J., {Yang}, A.Y., {Ganesh}, A., {Sastry}, S.S., {Ma}, Y.: Robust face
  recognition via sparse representation. IEEE Transactions on Pattern Analysis
  and Machine Intelligence  \textbf{31}(2),  210--227 (2009).
  \doi{10.1109/TPAMI.2008.79}

\bibitem{8920031}
{Yin}, H., {Wu}, X., {Chen}, S.: Locality constraint dictionary learning with
  support vector for pattern classification. IEEE Access  \textbf{7},
  175071--175082 (2019). \doi{10.1109/ACCESS.2019.2957417}

\bibitem{NIPS2009_2afe4567}
Yu, K., Zhang, T., Gong, Y.: Nonlinear learning using local coordinate coding.
  In: Bengio, Y., Schuurmans, D., Lafferty, J., Williams, C., Culotta, A.
  (eds.) Advances in Neural Information Processing Systems. vol.~22. Curran
  Associates, Inc. (2009),
  \url{https://proceedings.neurips.cc/paper/2009/file/2afe4567e1bf64d32a5527244d104cea-Paper.pdf}

\bibitem{5539989}
{Zhang}, Q., {Li}, B.: Discriminative k-svd for dictionary learning in face
  recognition. In: 2010 IEEE Computer Society Conference on Computer Vision and
  Pattern Recognition. pp. 2691--2698 (2010). \doi{10.1109/CVPR.2010.5539989}

\end{thebibliography}

\end{document}